\documentclass[sigconf,natbib=true,anonymous=false]{acmart}
\AtBeginDocument{%
  \providecommand\BibTeX{{%
    \normalfont B\kern-0.5em{\scshape i\kern-0.25em b}\kern-0.8em\TeX}}}

\usepackage{booktabs} 

\usepackage[english]{babel}
\usepackage{moresize}
\usepackage{amsmath}
\usepackage{algorithmic}
\usepackage{balance}
\usepackage{comment}
\usepackage{paralist}
\usepackage{bm}
\usepackage{pgfplots}
\usetikzlibrary{pgfplots.dateplot}

\usepackage{flushend}
\usepackage[english]{babel}
\usepackage[latin1]{inputenc}
\usepackage{mathrsfs}
\usepackage{graphicx}

\usepackage{amssymb}
\usepackage{amsfonts}
\usepackage{url}
\usepackage{longtable}
\usepackage{rotating}
\usepackage{multirow}
\usepackage{mathrsfs}
\usepackage{subfigure}
\usepackage{enumitem}
\usepackage[linesnumbered,algoruled,boxed,lined]{algorithm2e}
\usepackage{adjustbox}
\usepackage{hyperref}
\usepackage{pgfplots}
\usetikzlibrary{pgfplots.dateplot}
\usepackage{filecontents}
\usepackage{threeparttable}
\definecolor{tblue}{RGB}{31,119,180}
\definecolor{torange}{RGB}{255,127,14}
\definecolor{tgreen}{RGB}{44,160,44}
\definecolor{tred}{RGB}{214,39,40}
\definecolor{tpurple}{RGB}{148,103,189}
\usepackage{bbm}

\newcommand{\hide}[1]{} 

\newcommand{\eg}{\textit{e}.\textit{g}.} 
\newcommand{\wrt}{\textit{w}.\textit{r}.\textit{t}}

\def\model{CL4ST}

\copyrightyear{2023}
\acmYear{2023}
\setcopyright{acmlicensed}\acmConference[CIKM '23]{Proceedings of the 32nd ACM International Conference on Information and Knowledge Management}{October 21--25, 2023}{Birmingham, United Kingdom}
\acmBooktitle{Proceedings of the 32nd ACM International Conference on Information and Knowledge Management (CIKM '23), October 21--25, 2023, Birmingham, United Kingdom}
\acmPrice{15.00}
\acmDOI{10.1145/3583780.3615065}
\acmISBN{979-8-4007-0124-5/23/10}



\begin{document}

\title{Spatio-Temporal Meta Contrastive Learning}
\author{Jiabin Tang}
\affiliation{
  \institution{University of Hong Kong \country{China}}
}
\email{jiabintang77@gmail.com}

\author{Lianghao Xia}
\affiliation{
  \institution{University of Hong Kong \country{China}}
}
\email{aka_xia@foxmail.com}

\author{Jie Hu}
\affiliation{
  \institution{Southwest Jiaotong University \country{China}}
}
\email{jiehu@swjtu.edu.cn}

\author{Chao Huang}
\authornote{Chao Huang is the corresponding author.}
\affiliation{
  \institution{University of Hong Kong \country{China}}
}
\email{chaohuang75@gmail.com}
\renewcommand{\shortauthors}{Jiabin Tang, Lianghao Xia, Jie Hu, \& Chao Huang}

\begin{abstract}

Spatio-temporal prediction is crucial in numerous real-world applications, including traffic forecasting and crime prediction, which aim to improve public transportation and safety management. Many state-of-the-art models demonstrate the strong capability of spatio-temporal graph neural networks (STGNN) to capture complex spatio-temporal correlations. However, despite their effectiveness, existing approaches do not adequately address several key challenges. Data quality issues, such as data scarcity and sparsity, lead to data noise and a lack of supervised signals, which significantly limit the performance of STGNN. Although recent STGNN models with contrastive learning aim to address these challenges, most of them use pre-defined augmentation strategies that heavily depend on manual design and cannot be customized for different Spatio-Temporal Graph (STG) scenarios. To tackle these challenges, we propose a new spatio-temporal contrastive learning (\model) framework to encode robust and generalizable STG representations via the STG augmentation paradigm. Specifically, we design the meta view generator to automatically construct node and edge augmentation views for each disentangled spatial and temporal graph in a data-driven manner. The meta view generator employs meta networks with parameterized generative model to customize the augmentations for each input. This personalizes the augmentation strategies for every STG and endows the learning framework with spatio-temporal-aware information. Additionally, we integrate a unified spatio-temporal graph attention network with the proposed meta view generator and two-branch graph contrastive learning paradigms. Extensive experiments demonstrate that our \model\ significantly improves performance over various state-of-the-art baselines in traffic and crime prediction. Our model implementation is available at the 
link: \url{https://github.com/HKUDS/CL4ST}.

\end{abstract}




\maketitle

\section{Introduction}
\label{sec:intro}

Spatio-temporal prediction, with its focus on analyzing and extracting insights from large and diverse spatio-temporal datasets, has become increasingly vital in numerous real-world applications. Examples include traffic prediction~\cite{DCRNN, STGCN, zhang2023automated}, crime prediction~\cite{DeepCrime, ST-SHN}, and epidemic forecasting~\cite{covid19,CausalGNN}. By leveraging these predictive techniques, various challenging problems such as transportation management and public safety risk assessment can be addressed and alleviated effectively. At the heart of spatio-temporal prediction lies the ability to capture and understand the spatial and temporal correlations present in historical observations.


The advent of deep learning techniques has enabled significant progress in a range of spatio-temporal prediction tasks. For example, in traffic prediction, researchers have proposed models equipped with Recurrent Neural Networks (RNN)~\cite{DCRNN, STMGCN, DMVST-Net, AGCRN, STDN, ST-MetaNet} and Temporal Convolutional Networks (TCN)~\cite{GraphWaveNet, DMSTGCN, MTGNN, ASTGCN, STGODE} have been proposed to capture temporal variation patterns. In addition, Graph Neural Networks (GNN)\cite{STGCN, STMGCN, STGODE} and Convolutional Neural Networks (CNN)\cite{ST-ResNet, DMVST-Net, ConvLSTM, UrbanFM} are adopted to learn underlying spatial correlations. The self-attention mechanism has also been employed and shown to be effective in modeling spatio-temporal dependency~\cite{ST-WA, GMAN, STGSP}. On the other hand, in the context of crime prediction, recurrent attentive networks are utilized to model complicated spatio-temporal crime patterns~\cite{DeepCrime}, while Hypergraph Neural Networks~\cite{ST-SHN} and Self-Supervised Learning~\cite{ST-HSL} have been employed to learn global spatio-temporal dependencies and address specific challenges in learning crime patterns. \\\vspace{-0.12in}.

\noindent \textbf{Dilemmas.}
Despite the effectiveness of the above models in achieving state-of-the-art spatio-temporal prediction performance, there are still several key challenges that need to be addressed in order to further improve the accuracy and applicability of these models. \\\vspace{-0.12in}

\noindent \textbf{Data Quality Issues}. Real-world datasets used in spatio-temporal prediction tasks often suffer from data quality issues that cannot be ignored. These issues can be broadly categorized into two classes. \textbf{i)} \emph{Data Scarcity}: Public datasets frequently utilized in spatio-temporal prediction tasks often have a limited number of samples. For example, the PEMS-04 dataset~\cite{STSGCN} used in traffic prediction contains only 16,992 samples in total. In addition to the limited number of samples, data missing problems often occur in real-world spatio-temporal applications due to various reasons, such as sensor failure in traffic scenarios and data privacy in epidemic forecasting. \textbf{ii)} \emph{Data Sparsity} is another issue in some spatio-temporal forecasting tasks, such as crime prediction~\cite{ST-HSL} and epidemic forecasting~\cite{CausalGNN}. In these cases, the data of each fine-grained region or sensor can be sparse along the temporal dimension when compared to the whole urban space. This can result in a lack of supervision signals, making it challenging to accurately predict future trends. \\\vspace{-0.12in}


\noindent \textbf{Limited Augmentation Strategies}:
Several spatio-temporal approaches based on contrastive learning have recently been proposed to address issues related to data sparsity or data scarcity~\cite{STGCL,ST-HSL}. However, the augmentation strategies used in these models, which are a significant component of contrastive learning, are often manual and pre-defined. As a result, the effectiveness of these augmentation strategies can be highly dependent on the pre-defined strategies and cannot be customized for different time spans or regions. This makes the models less generalized and robust in real-world scenarios, where the spatio-temporal context can vary significantly. \\\vspace{-0.12in}

\noindent \textbf{Contribution.} 
To address the challenges outlined above, we propose a novel Spatio-Temporal Contrastive Learning (\model) framework that enhances the robustness and generalization of spatio-temporal graph neural networks by endowing them with self-supervised data augmentation. Our approach integrates a parameterized view generator with meta networks to automatically provide each graph with customized augmented node and edge views. This approach enables the meta view generator to obtain customized data augmentations to boost the effectiveness of contrastive learning and inject the extracted spatio-temporal information into the entire contrastive learning procedure. This work makes several key contributions, which are summarized as follows: \vspace{-0.05in}
\begin{itemize}[leftmargin=*]
\item In this work, we propose a new spatio-temporal meta contrastive learning framework, called \model, to strengthen the robustness and generalization capacity of spatio-temporal modeling. \\\vspace{-0.12in}

\item In our \model, the meta view generator automatically customizes node- and edge-wise augmentation views for each spatio-temporal graph according to the meta-knowledge of the input graph structure. This approach not only obtains personalized augmentations for every graph but also injects spatio-temporal contextual information into the data augmentation framework. \\\vspace{-0.12in}

\item We conduct extensive experiments to evaluate the effectiveness of the \model\ on spatio-temporal prediction tasks, such as traffic forecasting and crime prediction. Comparisons over different datasets show that \model\ outperforms state-of-the-art baselines.

\end{itemize}

\vspace{-0.12in}
\section{Preliminaries}
\label{sec:pre}

\noindent \textbf{Spatio-Temporal Prediction}
involves predicting future spatio-temporal signals based on historical observations. In general, spatio-temporal prediction can be classified into two categories: i) Graph-based approach: It involves deploying a network of $N$ sensors to monitor a specific volume in an urban area. Each sensor is represented as a node $v_n$ in the network, which is constructed as a graph. ii) Grid-based approach: It involves partitioning a city into $N = I \times J$ disjoint geographical grids, where each grid represents a spatial region $r_n$. Spatio-temporal data is represented as a grid, where each cell in the grid represents a spatial region. \\\vspace{-0.12in}


\noindent \textbf{Spatio-Temporal Graph (STG)}.
We model both the aforementioned tasks using Spatio-Temporal Graphs (STGs), which are defined as $\mathcal{G}(\mathcal{V}, \mathcal{E}, A, \mathbf{X})$ where $\mathcal{V}$ denotes a set of nodes or regions, with $|\mathcal{V}| = N$, $\mathcal{E}$ is a set of edges, and $A\in \mathbb{R}^{N\times N}$ represents an adjacency matrix. The feature matrix $\mathbf{X} \in \mathbb{R}^{T\times N\times F}$ is defined over the STG, and represents the matrix consisting of target attributes such as traffic volumes or crime records. Here, $F$ denotes the feature dimension and $T$ represents the number of time steps. \\\vspace{-0.12in}


\noindent \textbf{Problem Statement}. 
The aim of STG forecasting is to learn a function $f$ that can predict the specific volume of an STG in the next $T^{\prime}$ steps, based on $T$ historical frames.
\begin{align}
    \mathcal{G}(\mathcal{V}, \mathcal{E}, A, \mathbf{X}_{t-T:t-1}) \stackrel{f}{\longrightarrow} \mathcal{G}^{\prime}(\mathcal{V}, \mathcal{E}, A, \mathbf{Y}_{t:t+T^{\prime}-1}) \nonumber
\end{align}
where the observations are represented by the feature matrix $\mathbf{X}\in \mathbb{R}^{T\times N \times F}$, where $T$ is the number of time steps and $N$ is the number of regions or nodes in the STG. The matrix $\mathbf{X}$ contains the observations with $F$ features from the time step $t-T$ to $t-1$.

\vspace{-0.05in}
\section{Methodology}
\label{sec:solution}
 
\begin{figure}
    \centering
    \includegraphics[width=0.47\textwidth]{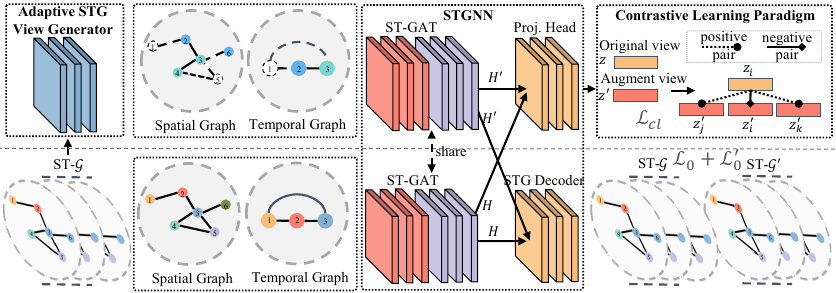}
    \vspace{-0.05in}
    \caption{Overall Framework of \model\ Model.}
    \vspace{-0.2in}
    \label{fig:overall}
\end{figure}

\begin{figure*}
    \centering
    \includegraphics[width=1\textwidth]{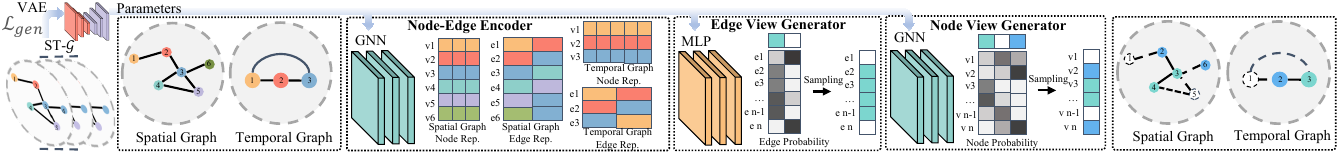}
    \vspace{-0.15in}
    \caption{Workflow of the meta view generator. The spatial (or temporal) graph signals are encoded by a GNN based on the original graph structure, resulting in STG graph node- and edge-wise representations. Then, node- and edge-wise augmented views are generated by the node and edge view generators, respectively. Parameters of encoders are obtained by VAE.}
    \vspace{-0.15in}
    \label{fig:meta_gen}
\end{figure*}

In this section, we present our \model\ framework and illustrate the overall model architecture in Figure~\ref{fig:overall}. The \model\ framework embeds both the original and augmented views of the STG, using a shared STG encoder to obtain the original and augmented STG representations, respectively. Additionally, the augmented view, as illustrated in Figure~\ref{fig:meta_gen}, is adaptively generated by the meta view generator, whose parameters are learned from the STG using the Variational Autoencoder (VAE). For training, we employ the contrastive learning paradigm with two branches and introduce the auxiliary loss from the VAE to control the learned parameters.

\vspace{-0.05in}
\subsection{Meta View Generator}


Previous studies~\cite{GCLA, ren2023disentangled} have demonstrated the crucial role of data augmentation with contrastive learning in graph representations. While recent works~\cite{JOAO, AutoGCL, ADGCL} have proposed adaptive approaches on graphs to automatically obtain task-dependent augmentation choices, there is still a research gap in existing methods to enable customized contrastive learning for spatio-temporal modeling.


In our model, we propose a meta view generator in our \model\ framework. The generator can learn the augmentation view in an automated way, utilizing the meta-knowledge from the input spatio-temporal graph. We argue that the designed generator can enhance the contrastive augmentation and thus obtain more robust and generalizable STG embeddings. Additionally, we inject spatio-temporal contextual information into the view generator, which can reflect spatio-temporal dependencies.


To demonstrate the effectiveness of our meta view generator, we analyze it on a graph $\mathcal{G}(\mathcal{V}, \mathcal{E}, A, \mathbf{X})$, where $A\in \mathbb{R}^{N\times N}$ is the adjacency matrix and $\mathbf{X}= {\Vec{x}_1, \Vec{x}_2, \cdots, \Vec{x}_N}, \Vec{x}_i \in \mathbb{R}^{f}$ denotes graph signals. We elaborate on the process of learnable view generation and meta networks. We elaborate on the process of learnable view generation and meta networks to showcase how our approach can automatically learn task-dependent augmentation choices and capture complex spatio-temporal dependencies.

\subsubsection{\bf Learnable View Generation}
Inspired by~\cite{AutoGCL}, our goal for learnable view generation is to design an end-to-end differentiable framework that can learn an augmented view on the graph $\mathcal{G}$. Specifically, the augmented view in our \model\ consists of a node view $f_{v}$ and an edge view $f_{e}$, which apply augmented strategies on the nodes and edges of the graph $\mathcal{G}$, respectively. For the node view $f_{v}$, we offer three different augmented operators: drop (drop nodes), keep (keep nodes unchanged), and mask (replace nodes with the mean value). We employ the Graph Isomorphism Network (GIN)\cite{GIN} to embed highly extracted graph representations over the graph $\mathcal{G}$ and utilize the Gumbel-Softmax reparametrization trick\cite{softmax_1, softmax_2} to enable differentiable sampling. This can be formalized as:
\begin{align}
    \vec{h}^{(1)}_{v} &= \mathcal{F}_{\Theta_1} [(1 + \epsilon^{(1)})\vec{h}^{(0)}_{v} + \sum_{u\in \mathcal{N}_{v}}\vec{h}^{(0)}_{u}] \nonumber \\
    \vec{h}^{(0)}_{v} &= \vec{x}^{(0)}_{v} , \vec{h}^{(0)}_{u} = \vec{x}^{(0)}_{u} \nonumber \\
    \vec{h}^{(2)}_{v} &= \mathcal{F}_{\Theta_2} [(1 + \epsilon^{(2)})\vec{h}^{(1)}_{v} + \sum_{u\in \mathcal{N}_{v}}\vec{h}^{(1)}_{u}] \nonumber \\
    f_{v} &= \text{GumbelSoftmax}(\vec{h}^{(2)}_{v})\label{eq:nodeview}
\end{align}
where $\mathcal{F}_{\Theta_1}$ and $\mathcal{F}_{\Theta_2}$ indicate MLPs with the parameters $\Theta_1$ and $\Theta_2$, respectively. $\vec{h}^{(1)}_{v}\in \mathbb{R}^{d_1}$ represents the extracted graph embeddings, while $\epsilon^{(1)}$ and $\epsilon^{(2)}$ are fixed scalars. $\vec{h}^{(2)}_{v}\in \mathbb{R}^{d_2}$ denotes the probabilities of choosing different augmentations, where $d_2 = 3$.
For the edge view $f_{e}$, we adopt two augmented strategies: drop (drop edges) and keep (keep edges unchanged). The procedure for the edge view generator can be formalized as follows:
\begin{align}
    \vec{h}^{(1)}_{e} &= \vec{h}^{(1)}_{v} \| \vec{h}^{(1)}_{u}, \mathrm{s.t.}, u\in\mathcal{N}_v  \nonumber \\
    \vec{h}^{(2)}_{e} &= \mathcal{F}_{\Theta_3} [\vec{h}^{(1)}_{e}] \nonumber \\
    f_{e} &= \text{GumbelSoftmax}(\vec{h}^{(2)}_{e})\label{eq:edgeview}
\end{align}
where $|$ indicates concatenation, $\vec{h}^{(1)}{e}\in \mathbb{R}^{2d_1}$ represents the edge representations, $\mathcal{F}{\Theta_3}$ is an MLP with parameters $\Theta_3$, and $\vec{h}^{(2)}_{e}$ denotes the probability of the two view augmentations. In particular, the original graph $\mathcal{G}(\mathcal{V}, \mathcal{E}, A, \mathbf{X})$ is first embedded with GIN and augmented by the edge view $f_{e}$, and then augmented by the node view $f_{v}$. This process can be defined as follows:
\begin{align}
    \mathcal{G}(\mathcal{V}^{\prime}, \mathcal{E}^{\prime}, A^{\prime}, \mathbf{X}) &= \text{Augm}(\mathcal{G}(\mathcal{V}, \mathcal{E}, A, \mathbf{X}), f_{e}) \nonumber\\
    \mathcal{G}(\mathcal{V}^{\prime}, \mathcal{E}^{\prime}, A^{\prime}, \mathbf{X}^{\prime}) &= \text{Augm}(\mathcal{G}(\mathcal{V}^{\prime}, \mathcal{E}^{\prime}, A^{\prime}, \mathbf{X}), f_{v}) \label{eq:aug}
\end{align}
where The symbol $\text{Augm}(\cdot, \cdot)$ represents the application of a specific augmented strategy (the latter) on a specific graph (the former).

\vspace{-0.05in}
\subsubsection{\bf Meta Networks for Generator}
To enhance the model customization ability with different augmented views, we use meta networks for the generator. Building on the motivation of previous work~\cite{ST-WA}, we introduce VAE~\cite{VAE} into the process of parameter generation, which is believed to have better generalization and representational power. Formally, the goal of parameter generation is to learn parameters $\Theta_i, i=1,2,3$ in Equations~\ref{eq:nodeview} and \ref{eq:edgeview} utilizing the graph features $\mathbf{X}$. This can be defined as follows:


\begin{align}
    \mathbf{z}^{(i)} \sim \mathcal{N}(\mu^{(i)}, \Sigma^{(i)});~~~
    \mathbf{z}^{(i)}_\phi \sim \mathcal{N}(\mu^{(i)}_\phi, \Sigma^{(i)}_\phi) \nonumber \\
    \mu^{(i)}_\phi, \Sigma^{(i)}_\phi = \mathcal{F}_\Phi[\mathbf{X}];~~~
    \Theta_{i} = \mathcal{F}_\Psi[\mathbf{z}^{(i)} + \mathbf{z}^{(i)}_\phi] \label{eq:paragen}
\end{align}
\noindent $\mu^{(i)}$ and $\Sigma^{(i)}$ are the learnable mean and covariance matrix, while $\mu^{(i)}\phi$ and $\Sigma^{(i)}\phi$ denote the mean and covariance matrix learned from the features $\mathbf{X}$ by an MLP $\mathcal{F}\Phi$ with learnable parameters $\Phi$. $\mathcal{F}_\Psi$ is an MLP with parameters $\Psi$. Due to the lack of prior knowledge of the latent space, we use Gaussian distributions that are well accepted by many previous works~\cite{ST-WA, VAE}, and the KL divergence to constrain the latent variables. This results in the following:
\begin{align}
    \mathcal{L}_{\text{gen}} &= \ensuremath{D_{KL}[(\mathbf{z}^{(i)} + \mathbf{z}^{(i)}_\phi)\| \hat{p}]};~~~
    \hat{p} &\sim \mathcal{N}(0, \mathbf{I}) \label{eq:gen_loss}
\end{align}
$\hat{p}$ represents a sample from the prior Gaussian distribution. It is worth noting that, according to empirical results, the generation of $\Theta_1$ and $\Theta_2$ follows the procedure independently in Equation~\ref{eq:paragen}, while $\Theta_3$ shares the value with $\Theta_2$ in practical implementations.

\vspace{-0.05in}
\subsection{Spatio-Temporal Graph Attention Networks}
Graph Neural Networks (GNNs) have become a popular and powerful tool for capturing complex correlations, particularly in spatio-temporal mining~\cite{graphAE, GIN, GAT, DCRNN, STGCN}. To fully exploit the advantages of GNNs, we employ a unified GNN encoder inspired by Graph Attention Networks (GAT)\cite{GAT} to reason about spatio-temporal dynamics. Following the approach in\cite{STFGNN}, we use a unified GNN-based framework to capture spatio-temporal dependencies on a unified spatio-temporal graph structure $\mathbf{A} \in \mathbb{R}^{TN \times TN}$. To avoid the enormous time complexity, we decouple the unified spatio-temporal graph into a temporal graph and a spatial graph during the modeling process. To begin with, we embed the STG feature matrix $\mathbf{X} \in \mathbb{R}^{T \times N \times F}$ into a $d$-dimensional latent space using a linear transformation:
\begin{align}
    \mathbf{X}^{(0)} &= \mathbf{W}^{(0)} \cdot \mathbf{X} + \mathbf{b}^{(0)} \label{eq:start_fc}
\end{align}
where $\mathbf{X}^{(0)} \in \mathbb{R}^{T \times N \times d}$ represents the initialized STG embeddings, and $\mathbf{W}^{(0)} \in \mathbb{R}^{d \times F}$ and $\mathbf{b}^{(0)} \in \mathbb{R}^{d}$ are the weight and bias parameters. To encode spatial and temporal patterns using graph attention networks, we further embed $\mathbf{X}^{(0)}$ with a fully connected layer:
\begin{align}
    \mathbf{X}^{(s)} &= \mathbf{W}^{(s)} \cdot [\text{reshape}(\mathbf{X}^{(0)})] + \mathbf{b}^{(s)} \label{eq:fcs}
\end{align}
where $\mathbf{W}^{(s)}\in \mathbb{R}^{d^{(s)} \times (T*d)}$ and $\mathbf{b}^{(s)}\in \mathbb{R}^{d^{(s)}}$ are weight and bias matrices.
$\mathbf{X}^{(s)} = \{\Vec{x}_1^{(s)}, \Vec{x}_2^{(s)}, \cdots, \Vec{x}_N^{(s)}\}, \Vec{x}_i^{(s)}\in \mathbb{R}^{d^{(s)}}$ denotes spatial features. We extend the aforementioned definition of the STG $\mathcal{G}$ to a spatial graph $\mathcal{G}^{(s)}(\mathcal{V}^{(s)}, \mathcal{E}^{(s)}, A^{(s)}, \mathbf{X}^{(s)})$, 
where $A^{(s)}\in \mathbb{R}^{N\times N}$ indicates the spatial adjacency matrix. 
With the spatial graph, graph attention networks equipped with stacked multiple multi-head graph attention layers aim to capture spatial correlations. The graph attention layer is defined as follows:
\begin{align}
    \Vec{h}^{(s)}_i &= \mathop{\Bigm|\Bigm|}\limits_{k=1}^K\sum_{j\in \mathcal{N}_i \cup \{i\}} \alpha_{ij}^{k} \mathbf{W}^{k} \Vec{x}^{(s)}_i \nonumber \\ 
    \alpha_{ij} &= \frac{\exp(\text{LeakyReLU}(\vec{a}^\top[\mathbf{W} \Vec{x}^{(s)}_i + \mathbf{W} \Vec{x}^{(s)}_j]))}{\sum_{k\in \mathcal{N}_i \cup \{i\}}\exp(\text{LeakyReLU}(\vec{a}^\top[\mathbf{W} \Vec{x}^{(s)}_i + \mathbf{W} \Vec{x}^{(s)}_k]))} \label{eq:gat}
\end{align}
$\Vert$ indicates concatenation, $\mathcal{N}_i$ represents the set of neighbors of the $i^{th}$ node defined by $A^{(s)}$, $K$ is the number of heads, $\mathbf{W} \in \mathbb{R}^{d^{(s)} \times d^{(s)}}$ represents the weight matrix, and $\vec{a} \in \mathbb{R}^{d^{(s)}}$ denotes the weight vector. After passing through the stacked GAT layers, we obtain the extracted spatial embeddings $\mathbf{H}^{(s)} = {\Vec{h}_1^{(s)}, \Vec{h}_2^{(s)}, \cdots, \Vec{h}_N^{(s)}}, \Vec{h}_i^{(s)} \in \mathbb{R}^{d^{(s)}}$. Next, the spatial embeddings $\mathbf{H}^{(s)}$ are transformed into the feature matrix $\mathbf{H}^{\prime (s)} \in \mathbb{R}^{T \times N \times d}$ using a linear layer as follows:
\begin{align}
    \mathbf{H}^{\prime (s)} &= \text{reshape}[\mathbf{W}^{(1)} \cdot [\text{reshape}(\mathbf{H}^{(s)})] + \mathbf{b}^{(1)}] \label{eq:final}
\end{align}
where $\mathbf{W}^{(1)}\in \mathbb{R}^{(T*d) \times d^{(s)}}$ and $\mathbf{b}^{(s)}\in \mathbb{R}^{(T*d)}$ are weight and bias matrices. 
As for the temporal graph, we employ a similar definition to the spatial, that is $\mathcal{G}^{(t)}(\mathcal{V}^{(t)}, \mathcal{E}^{(t)}, A^{(t)}, \mathbf{X}^{(t)})$, in which $A^{(t)}\in \mathbb{R}^{T\times T}$ denotes the temporal adjacency matrix expressing the correlations among different time steps, $\mathbf{X}^{(t)}= \{\Vec{x}_1^{(t)}, \Vec{x}_2^{(t)}, \cdots, \Vec{x}_T^{(t)}\}, \Vec{x}_i^{(t)}\in \mathbb{R}^{d^{(t)}}$ represents the temporal feature matrix. In particular, $\mathbf{X}^{(t)}$ is generated from $\mathbf{H}^{\prime (s)}$ using a similar fully connected layer as in Equation~\ref{eq:fcs}. To further capture the temporal dependencies, we adopt the stacked multi-head GAT layers formalized analogously to Equation~\ref{eq:gat}, resulting in the temporal features with the definition of $\mathbf{H}^{(t)} = \{\Vec{h}_1^{(t)}, \Vec{h}_2^{(t)}, \cdots, \Vec{h}_T^{(t)}\}, \Vec{h}_i^{(t)}\in \mathbb{R}^{d^{(t)}}$.


Ultimately, we convert the temporal feature matrix $\mathbf{H}^{(t)}$ into the final feature matrix $\mathbf{H} = \mathbf{H}^{\prime (t)} \in \mathbb{R}^{T \times N \times d}$ using a similar function as in Equation~\ref{eq:final}. To summarize how to construct spatial and temporal graphs: (i) \textbf{Spatial graph ($A^{(s)}$):} The spatial graph represents the correlations between spatial units. For the two common types of spatio-temporal prediction, graph-based and grid-based~\cite{DL-Traff}, we can construct graphs using a thresholded Gaussian kernel~\cite{DCRNN} and considering neighboring regions as neighbors~\cite{ST-HSL, ST-SHN}, respectively. (ii) \textbf{Temporal graph ($A^{(t)}$):} The temporal graph represents the correlations between temporal representations at different time steps. Formally, if the historical time step is $T$, we have the temporal graph $A^{(t)} \in \mathbb{R}^{T \times T}$, and $A^{(t)}_{i,j} = 1$ for arbitrary $i, j$. This means that we assume that every time step influences others originally. Applying the GAT network for information propagation on the temporal graph is equivalent to existing works (\eg~\cite{GMAN}) that utilize the self-attention mechanism to capture temporal correlations.

\vspace{-0.05in}
\subsection{Spatio-Temporal Graph Decoder Layer.}
\label{sec:decoder}
With the final features $\mathbf{H}$ learned by the foregoing spatio-temporal graph attention networks, we can design a spatio-temporal graph decoder layer to construct the predictive results.

\vspace{-0.05in}
\subsubsection{\bf Spatio-Temporal Position-Aware Encoding}
To enhance the model capacity in identifying different spatial and temporal positions (nodes and time steps, respectively), we adopt ideas from~\cite{STID} and introduce learnable spatial position $E^{(s)} \in \mathbb{R}^{N \times D}$ and temporal positions, which are composed of \emph{time of day} embeddings $E^{(\text{TiD})} \in \mathbb{R}^{T \times D}$ and \emph{day of week} embeddings $E^{(\text{DiW})} \in \mathbb{R}^{T \times D}$~\cite{GraphWaveNet}. For implementation, we randomly initialize a tensor $E^{(s)} \in \mathbb{R}^{N \times D}$, and the value of the tensor can be updated during backpropagation. As for temporal positional embeddings, we randomly initialize a \emph{time of day} tensor $E^{(\text{TiD})}{\text{all}} \in \mathbb{R}^{288 \times D}$ and a \emph{day of week} tensor $E^{(\text{DiW})}{\text{all}} \in \mathbb{R}^{7 \times D}$, where 288 denotes that a day has 288 time steps, and 7 denotes that a week has 7 days. The input \emph{time of day} and \emph{day of week} indices of the STG query the \emph{time of day} and \emph{day of week} tensors to obtain temporal positional embeddings.


\vspace{-0.05in}
\subsubsection{\bf Information Fusion}
Eventually, we employ the concatenation operation (denoted by $|$) to integrate the final feature matrix $\mathbf{H}$, the spatial and temporal positions ($E^{(s)}$, $E^{(\text{TiD})}$, and $E^{(\text{DiW})}$), and the initialized STG embeddings $\mathbf{X}^{(0)} \in \mathbb{R}^{T\times N\times d}$ for residual connection, which is formalized as follows:
\begin{align}
    \mathbf{Y} &=  \mathcal{F}_{\Omega_2}[\mathcal{F}_{\Omega_1}(\mathbf{H}) \| E^{(s)} \| E^{(\text{TiD})} \| E^{(\text{DiW})} \| \mathcal{F}_{\Omega_1}(\mathbf{X}^{(0)})] \label{eq:pred_layer}
\end{align}
Here, $\mathcal{F}_{\Omega_1}$ and $\mathcal{F}_{\Omega_2}$ refer to MLP networks with parameter sets $\Omega_1$ and $\Omega_2$, respectively. $\mathbf{Y} \in \mathbb{R}^{T^{\prime}\times N \times F^{\prime}}$ indicates the prediction.

\vspace{-0.05in}
\subsection{Contrastive Learning Paradigm}
After elaborating on the three key components above, we present the entire workflow and GCL paradigm in our model. Overall, there are two branches in the proposed model, namely the original branch and the augmented branch. In the original branch, $\mathcal{G}(\mathcal{V}, \mathcal{E}, A, \mathbf{X})$ is regarded as the spatial graph $\mathcal{G}^{(s)}(\mathcal{V}^{(s)}, \mathcal{E}^{(s)}, A^{(s)}, \mathbf{X}^{(s)})$ and the temporal graph $\mathcal{G}^{(t)}(\mathcal{V}^{(t)}, \mathcal{E}^{(t)}, A^{(t)}, \mathbf{X}^{(t)})$, and is fed into the aforementioned spatio-temporal GAT to obtain the original STG representations $\mathbf{H}$. In the augmented branch, we inject the augmentations into the STG utilizing spatial and temporal view generators in Equation~\Ref{eq:aug} with meta-parameters in Equation~\Ref{eq:paragen} and embed the augmented STG with the shared spatio-temporal Graph Attention Networks to obtain the augmented STG representations $\mathbf{H}^{\prime}$. Regarding contrastive learning, we adopt the graph-level contrast following~\cite{AutoGCL, STGCL}, which has been proven to be effective in STG forecasting tasks. Specifically, we employ the projection head to map the STG representations $\mathbf{H}$ and $\mathbf{H}^{\prime}$ from both branches into the high-dimensional vector space and obtain representations $\vec{z}$ and $\vec{z}^{\prime}\in \mathbb{R}^{d^{(z)}}$ with fully connected layers.

Assuming there are $B$ STGs in a data batch, we consider two different views from the same input STG as the positive view pair, and otherwise as negative view pairs. Hence, we have:
\begin{align}
    \ell(\vec{z}_i, \vec{z}_i^{\prime}) = -\log(\frac{\exp(sim(\vec{z}_i, \vec{z}_i^{\prime})/\tau )}{\sum^{B}_{k=1}\mathbbm{1}_{[j\neq i]}\exp(sim(\vec{z}_i, \vec{z}^{\prime}_j)/\tau )}) \nonumber \\
    sim(\vec{z}_i, \vec{z}_j) = \frac{\vec{z}_i \cdot \vec{z}_j}{\left\lVert \vec{z}_i\right\rVert \cdot \left\lVert \vec{z}_j\right\rVert };~~~
    \mathcal{L}_{\text{cl}} = \frac{1}{B} \sum_{i=1}^{B}[\ell(\vec{z}_i, \vec{z}_i^{\prime}) + \ell(\vec{z}^{\prime}_i, \vec{z}_i) ] \label{eq:cl}
\end{align}
$\mathbbm{1}{[j\neq i]}\in {0, 1}$ denotes the indicator function for contrastive pairs, $\ell(\cdot)$ is the contrastive function for the given pair, and $\mathcal{L}_{cl}$ represents the contrastive loss for the whole batch of data.

\vspace{-0.05in}
\subsection{Model Optimization}
In this subsection, we discuss the learning process of the proposed \model. Primarily, the original and augmented STG representations $H$ and $H^{\prime}$ are fed into the predictive layers of the spatio-temporal graph decoder layer in Subsection~\Ref{sec:decoder}, resulting in predictive results $\hat{\mathbf{Y}}$ and $\hat{\mathbf{Y}}^{\prime}\in \mathbb{R}^{T^{\prime}\times N\times F^{\prime}}$. We then calculate the predictive loss as:
\begin{align}
    \mathcal{L}_{\text{pre}} = \ell(\mathbf{Y}, \hat{\mathbf{Y}}) + \ell(\mathbf{Y}, \hat{\mathbf{Y}}^{\prime})  \label{eq:pred_loss}
\end{align}
Here, $\mathbf{Y}\in \mathbb{R}^{T^{\prime}\times N\times F^{\prime}}$ represents the ground-truth STG signals, and $\ell(\cdot, \cdot)$ is the specific loss function that varies from task to task. For instance, in our experiments, we employ the Huber loss~\cite{huber1992robust} for the traffic forecasting task, which is defined as follows:
\begin{equation}
    \label{eq6}
    \ell(\mathbf{Y}, \hat{\mathbf{Y}}) = \mathcal{H}(\mathbf{Y}, \hat{\mathbf{Y}})=\left\{
    \begin{aligned}
    &\frac{1}{2}(\mathbf{Y} - \hat{\mathbf{Y}})  , & \left\lvert \mathbf{Y} - \hat{\mathbf{Y}} \right\rvert  \leq  \delta  \\
    &\delta(\left\lvert \mathbf{Y} - \hat{\mathbf{Y}} \right\rvert - \frac{1}{2}\delta)  , & otherwise
    \end{aligned}
    \right.
\end{equation}
$\delta$ denotes a threshold value, while as to the crime prediction task, we follow the mean absolute error (MSE) loss~\cite{ST-HSL} and have 
\begin{align}
    \ell(\mathbf{Y}, \hat{\mathbf{Y}}) = \left\lVert \mathbf{Y} - \hat{\mathbf{Y}} \right\rVert^2_2
\end{align}
The joint loss of our \model\ in the training process is defined as:
\begin{align}
    \mathcal{L} = \mathcal{L}_{\text{pre}} + \lambda_1\mathcal{L}_{\text{cl}} + \lambda_2\mathcal{L}_{\text{s-gen}} + \lambda_3\mathcal{L}_{\text{t-gen}} \label{eq:loss_final}
\end{align} 
where $\mathcal{L}_{\text{cl}}$ indicates the contrastive loss in Equation~\Ref{eq:cl}, $\lambda_i, i=1,2,3$ are coefficients for controlling the loss, and $\mathcal{L}_{\text{s-gen}}$ and $\mathcal{L}_{\text{t-gen}}$ represent the KL divergence loss in Equation~\Ref{eq:gen_loss} for the spatial and temporal meta view generators, respectively.

\vspace{-0.05in}
\section{Evaluation}
\label{sec:exp}
To evaluate the performance of \model, we conduct extensive experiments on three real-world traffic datasets and two crime datasets by answering the following research questions:
\begin{itemize}[leftmargin=*]
  \item \textbf{RQ1}: How does \model\ perform compared to SOTA prediction baselines while predicting future traffic volumes and crimes?
  \item \textbf{RQ2}: How do the key components contribute to the predictive performance of the \model\ framework?
  \item \textbf{RQ3}: How good is the generalization and robustness of \model?
  \item \textbf{RQ4}: How do various parameters influence model accuracy?
  \item \textbf{RQ5}: What is the model interpretation ability of our \model?
\end{itemize}

\vspace{-0.05in}
\subsection{Experimental Settings}
\subsubsection{\bf Datasets}
We conduct experiments on both citywide traffic prediction tasks and crime prediction tasks, utilizing five real-world datasets. The statistics of the datasets are shown in Table~\Ref{tb:ds}. We provide data detailed descriptions as follows:
\begin{table}
  \centering
  \caption{Statistical information of the experimental datasets.}
  \vspace{-0.15in}
  \resizebox{.47\textwidth}{!}{\begin{tabular}{ccccccc} 
  \hline
  Dataset   & Type  & Volume  & \# Interval & \# Nodes & \# Time Span      & \# Features  \\ 
\hline
PeMSD4    & Graph & Traffic & 5 min       & 307      & 01/2018 - 02/2018 & 1            \\
PeMSD7    & Graph & Traffic & 5 min       & 883      & 05/2017 - 08/2017 & 1            \\
PeMSD8    & Graph & Traffic & 5 min       & 170      & 07/2016 - 08/2016 & 1            \\
NYC Crime & Grid  & Crime   & 1 day       & 256      & 01/2014 - 12/2015 & 4            \\
CHI Crime & Grid  & Crime   & 1 day       & 168      & 01/2016 - 12/2017 & 4            \\
  \hline
  \end{tabular}}
  \label{tb:ds}
  \vspace{-0.1in}
\end{table}

\begin{itemize}[leftmargin=*]
  \item \textbf{Traffic Prediction}: 
  We utilize the PeMS04, PeMS07, and PeMS08 traffic datasets to evaluate the performance of our graph-based spatio-temporal modeling approach. These datasets are widely used in previous work~\cite{STGCN, ASTGCN, STSGCN, STGODE} and are collected by the California Performance of Transportation (PeMS)~\cite{chen2001freeway}, with a time interval of 5 minutes and different time spans.
  
  \item \textbf{Crime Prediction}:
  We also investigate the ability of our model to handle spatio-temporal prediction tasks on crime datasets, namely NYC Crime and CHI Crime~\cite{ST-HSL, ST-SHN}, which were collected from New York City (NYC) and Chicago, respectively, with a temporal resolution of 1 day. These datasets contain different crime types (e.g., robbery, larceny, etc.) and are generated using a spatial partition unit of $3$ km $\times$ $3$ km.
  
\end{itemize}

\subsubsection{\bf Evaluation Protocols}
In this subsection, we elaborate the details of our evaluation protocols as follows: \\\vspace{-0.12in}

\noindent\textbf{Traffic Prediction}: To conduct a fair comparison, we follow the dataset division used in previous studies~\cite{STGCN, ASTGCN, STSGCN, STGODE} and split the datasets into training, validation, and testing sets in a 6:2:2 ratio. \\\vspace{-0.12in}


\noindent\textbf{Crime Forecasting}: Following recent works~\cite{ST-HSL, ST-SHN}, we construct the training and testing sets with a ratio of 7:1, and we use crime records from the last month in the training set for validation. \\\vspace{-0.12in}
  
\noindent \textbf{Metrics}: We employ three widely used metrics, including \emph{Mean Absolute Error (MAE)}, \emph{Root Mean Squared Error (RMSE)}, and \emph{Mean Absolute Percentage Error (MAPE)}, for performance evaluation of both traffic and crime prediction.

  

\subsubsection{\bf Baseline Models}
For the traffic prediction evaluation, we utilize 18 baselines. On the other hand, for the crime forecasting evaluation, we compare \model\ with 12 baselines.

\begin{table*}[h]
  \centering
  \caption{Overall traffic forecasting performance on PeMSD4, 7, 8 in terms of MAE, RMSE, MAPE.}
  \vspace{-0.15in}
  \resizebox{1\textwidth}{!}{\begin{tabular}{c|c|cccccccccccccccccc|c} 
  \hline
  \multicolumn{2}{c|}{Model}         & HA      & VAR     & DCRNN   & STGCN   & DSANet  & GWN     & ASTGCN  & LSGCN   & STSGCN  & StemGNN & AGCRN   & STFGNN  & STGODE  & {\begin{tabular}[c]{@{}c@{}}Z-\\ GCNETs\end{tabular}} & {\begin{tabular}[c]{@{}c@{}}TAMP-\\ S2GCNets\end{tabular}} & FOGS    & GMSDR   & {\begin{tabular}[c]{@{}c@{}}STG-\\ NCDE\end{tabular}} & \textbf{\model}  \\ 
  \hline
  \multirow{3}{*}{\rotatebox{90}{PEMS4}} & MAE      & 38.03   & 24.54   & 21.22   & 21.16   & 22.79   & 24.89   & 22.93   & 21.53   & 21.19   & 21.61   & 19.83   & 19.83   & 20.84   & 19.50    & 19.74         & 19.74   & 20.49   & 19.21    & \textbf{18.49}    \\
                          & RMSE     & 59.24   & 38.61   & 33.44   & 34.89   & 35.77   & 39.66   & 35.22   & 33.86   & 33.65   & 33.80   & 32.26   & 31.88   & 32.82   & 31.61    & 31.74         & 31.66   & 32.13   & 31.09    & \textbf{30.17}    \\
                          & MAPE(\%) & 27.88 & 17.24 & 14.17 & 13.83 & 16.03 & 17.29 & 16.56 & 13.18 & 13.90 & 16.10 & 12.97 & 13.02 & 13.77 & 12.78  & 13.22       & 13.05 & 14.15 & 12.76  & \textbf{12.00}  \\ 
  \hline
  \multirow{3}{*}{\rotatebox{90}{PEMS7}} & MAE      & 45.12   & 50.22   & 25.22   & 25.33   & 31.36   & 26.39   & 24.01   & 27.31   & 24.26   & 22.23   & 22.37   & 22.07   & 22.99   & 21.77    & 21.84         & 21.28   & 22.27   & 20.53    & \textbf{20.20}    \\
                          & RMSE     & 65.64   & 75.63   & 38.61   & 39.34   & 49.11   & 41.50   & 37.87   & 41.46   & 39.03   & 36.46   & 36.55   & 35.80   & 37.54   & 35.17    & 35.42         & 34.88   & 34.94   & 33.84    & \textbf{34.06}    \\
                          & MAPE(\%) & 24.51 & 32.22 & 11.82 & 11.21 & 14.43 & 11.97 & 10.73 & 11.98 & 10.21 & 9.20  & 9.12  & 9.21  & 10.14 & 9.25   & 9.24        & 8.95  & 9.86  & 8.80   & \textbf{8.53}   \\ 
  \hline
  \multirow{3}{*}{\rotatebox{90}{PEMS8}} & MAE      & 34.86   & 19.19   & 16.82   & 17.50   & 17.14   & 18.28   & 18.25   & 17.73   & 17.13   & 15.91   & 15.95   & 16.64   & 16.81   & 15.76    & 16.36         & 15.73   & 16.36   & 15.45    & \textbf{14.74}    \\
                          & RMSE     & 52.04   & 29.81   & 26.36   & 27.09   & 26.96   & 30.05   & 28.06   & 26.76   & 26.80   & 25.44   & 25.22   & 26.22   & 25.97   & 25.11    & 25.98         & 24.92   & 25.58   & 24.81    & \textbf{24.17}    \\
                          & MAPE(\%) & 24.07 & 13.10 & 10.92 & 11.29 & 11.32 & 12.15 & 11.64 & 11.20 & 10.96 & 10.90 & 10.09 & 10.60 & 10.62 & 10.01  & 10.15       & 9.88  & 10.28 & 9.92   & \textbf{9.61}   \\
  \hline
  \end{tabular}}\label{tab:cmp1}%
  \vspace{-0.05in}
  \end{table*}

\begin{table}[h]
    \centering
    \caption{Overall performance comparison on NYC and CHI crime data in terms of MAE, RMSE, MAPE}
    \vspace{-0.1in}
    \resizebox{.47\textwidth}{!}{\begin{tabular}{cccccccc} 
    \hline
    \multirow{2}{*}{Model} & Dataset        & \multicolumn{3}{c}{NYC Crime}                       & \multicolumn{3}{c}{CHI Crime}                        \\ 
    \cline{2-8}
                           & Metrics        & MAE             & MAPE            & RMSE            & MAE             & MAPE            & RMSE             \\ 
    \hline
    \multicolumn{2}{c}{ARIMA}               & 1.0765          & 0.6196          & 1.5398          & 1.2616          & 0.5894          & 1.8398           \\
    \multicolumn{2}{c}{SVM}                 & 1.2805          & 0.6863          & 1.9216          & 1.3622          & 0.5992          & 2.0671           \\
    \multicolumn{2}{c}{ST-ResNet}           & 0.9755          & 0.5453          & 1.4065          & 1.1014          & 0.5294          & 1.6468           \\
    \multicolumn{2}{c}{DCRNN}               & 0.9638          & 0.5569          & 1.3730          & 1.0885          & 0.5260          & 1.5855           \\
    \multicolumn{2}{c}{STGCN}               & 0.9538          & 0.5451          & 1.3915          & 1.0970          & 0.5283          & 1.5845           \\
    \multicolumn{2}{c}{STtrans}             & 0.9640          & 0.5584          & 1.3755          & 1.0817          & 0.5179          & 1.5826           \\
    \multicolumn{2}{c}{DeepCrime}           & 0.9429          & 0.5496          & 1.3315          & 1.0801          & 0.5166          & 1.5636           \\
    \multicolumn{2}{c}{STDN}                & 0.9993          & 0.5762          & 1.3974          & 1.1245          & 0.5480          & 1.6470           \\
    \multicolumn{2}{c}{ST-MetaNet}          & 0.9572          & 0.5620          & 1.3462          & 1.0913          & 0.5225          & 1.5723           \\
    \multicolumn{2}{c}{GMAN}                & 0.9587          & 0.5575          & 1.3461          & 1.0752          & 0.5166          & 1.5515           \\
    \multicolumn{2}{c}{ST-SHN}              & 0.9280          & 0.5373          & 1.3168          & 1.0689          & 0.5116          & 1.5474           \\
    \multicolumn{2}{c}{DMSTGCN}             & 0.9293          & 0.5485          & 1.3167          & 1.0736          & 0.5175          & 1.5296           \\ 
    \hline
    \multicolumn{2}{c}{\textbf{\model}}   & \textbf{0.8819} & \textbf{0.5280} & \textbf{1.2892} & \textbf{1.0411} & \textbf{0.4981} & \textbf{1.5192}  \\
    \hline
    \end{tabular}}\label{tab:cmp2}
    \vspace{-0.2in}
    \end{table}
    
\noindent\textbf{Traffic Prediction:} 
\begin{itemize}[leftmargin=*]
  \item \textbf{HA}~\cite{HA}: This method integrates the moving average value of the observed time series to capture temporal dynamics.
  \item \textbf{VAR}~\cite{VAR}: A time series forecasting model that utilizes vector autoregression to predict traffic series of all nodes.
  \item \textbf{DCRNN}~\cite{DCRNN}: It utilizes a diffusional convolutional operation with a RNN model to model spatio-temporal correlations.
  \item \textbf{STGCN}~\cite{STGCN}: The model combines spatio-temporal graph convolutional networks with temporal gated convolutional networks.
  \item \textbf{DSANet}~\cite{DSANet}: It employs a dual self-attention to capture dynamic-periodic or nonperiodic patterns for multivariate signals.
  \item \textbf{GWN}~\cite{GraphWaveNet}: This framework integrates diffusional graph convolutions with an adaptive graph matrix into dilated 1D convolutions.
  \item \textbf{ASTGCN}~\cite{ASTGCN}: It injects attention mechanisms into spatio-temporal convolutional networks with three temporal properties of traffic flows to capture dynamic spatio-temporal dependencies.
  \item \textbf{LSGCN}~\cite{LSGCN}: It integrates graph convolution networks into gated linear units convolution for both long- and short-term prediction.
  \item \textbf{STSGCN}~\cite{STSGCN}: The model adopts a spatio-temporal synchronous modeling mechanism to capture spatio-temporal heterogeneities. 
  \item \textbf{StemGNN}~\cite{StemGNN}: It combines Graph Fourier Transform and Discrete Fourier Transform with 1D convolutional layers.
  \item \textbf{AGCRN}~\cite{AGCRN}: It uses graph convolutional recurrent networks with node adaptive parameter learning and data-adaptive graph generation modules to capture node-specific spatial patterns.
  \item \textbf{STFGNN}~\cite{STFGNN}: It proposes a fusion operation that combines different spatial and temporal graphs for spatio-temporal reasoning. 
  \item \textbf{STG-ODE}~\cite{STGODE}: It combines a tensor-based ordinary differential equation with a semantical adjacency matrix to capture spatio-temporal dynamics and semantic information synchronously.
  \item \textbf{Z-GCNETs}~\cite{Z-GCNETs}: It integrates the most salient time-conditioned topological information and the concept of zigzag persistence into time-aware graph convolutional networks.
  \item \textbf{TAMP-S2GCNets}~\cite{TAMP-S2GCNets}: The model introduces time-aware multipersistence into spatio-supra graph convolutional networks. 
  \item \textbf{FOGS}~\cite{FOGS}: It employs first-order gradients to learn correlation graphs and address irregularly-shaped data distribution issues.
  \item \textbf{GMSDR}~\cite{GMSDR}: It introduces a multi-step dependency relation into graph convolutional operations and recurrent neural networks for long-term temporal modeling.
  \item \textbf{STG-NCDE}~\cite{STGNCDE}: This work uses neural controlled differential equations to process spatio-temporal graph modeling for capturing the complex patterns in traffic data.
\end{itemize}

\noindent\textbf{Crime Prediction:} 
\begin{itemize}[leftmargin=*]
  \item \textbf{SVM}~\cite{SVR}: This model utilizes support vector machines to predict non-linear and non-stationary temporal patterns in traffic data.
  \item \textbf{ST-ResNet}~\cite{ST-ResNet}: It employs convolutional neural networks with residual connections and three temporal properties of traffic flows to capture spatio-temporal patterns.
  \item \textbf{STtrans}~\cite{STtrans}: It uses stacked transformer layers with query/key transformations to explore spatio-temporal sparse data.
  \item \textbf{DeepCrime}~\cite{DeepCrime}: This model integrates attention mechanisms into temporal recurrent neural networks for crime prediction.
  \item \textbf{STDN}~\cite{STDN}: A periodic shifted attention and flow gating scheme are used in this framework for dynamic similarity reasoning.
  \item \textbf{ST-MetaNet}~\cite{ST-MetaNet}: The meta-learning methods with graph-based sequence-to-sequence paradigm is used to extract diverse meta knowledge from spatio-temporal data. 
  \item \textbf{GMAN}~\cite{GMAN}: Spatio-temporal graph encoder and decoder with multi-attention networks is adopted in this work.
  \item \textbf{ST-SHN}~\cite{ST-SHN}: It employs hypergraph convolutional networks to encode spatial information among different geographical regions.
  \item \textbf{DMSTGCN}~\cite{DMSTGCN}: It integrates dynamic graph generator into multi-faceted spatio-temporal graph convolutional networks.
\end{itemize}

\vspace{-0.05in}
\subsubsection{\bf Implementation Details}


We implement our \model\ with PyTorch and the PyTorch Geometric library and adopt Adam as the optimizer for model training. We also utilize a batch size of 16 and schedule the initial learning rate at $1e^{-3}$ using a decay ratio of 0.5 with epoch steps [1, 50, 100]. As for the model hyperparameters, we employ two GAT layers with 4 heads for spatial encoding and 1 head for temporal encoding. The spatial dimension $d^{(s)}$ is set to 64, while the temporal dimension $d^{(t)}$ is set to 128. The dimension of the latent variables in Equation~\ref{eq:paragen} is set to 16. We adopt an annealing strategy to control $\lambda_1$, $\lambda_2$, and $\lambda_3$ in Equation~\ref{eq:loss_final}, gradually changing them from 0 to 1 as the epoch increases to balance the loss. For traffic forecasting, we consider a sequential length of 12 time steps of historical traffic records to predict the next 12 time steps of traffic volumes. This task can be described as a 12-sequence-to-12-sequence prediction. For crime prediction, we are predicting the next 1 day of crime data based on the past 30 days. More detailed implementation information can be found in our released code.

\begin{figure}
  \centering
  \subfigure[node 181, from time step 0$\sim$288]{
    \centering
    \includegraphics[width=0.18\textwidth]{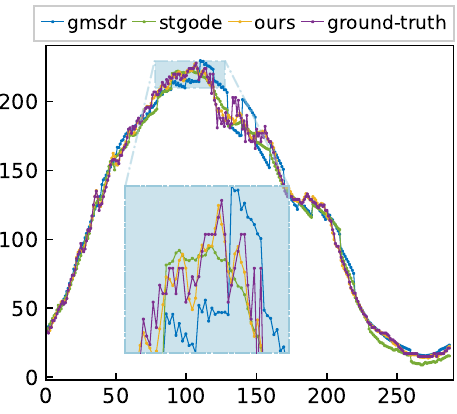}
}
\subfigure[node 181, from time step 2505$\sim$2793]{
    \centering
    \includegraphics[width=0.18\textwidth]{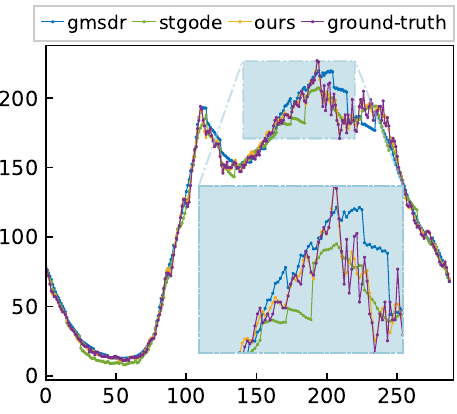}
}
\subfigure[node 198, from time step 0$\sim$288]{
    \centering
    \includegraphics[width=0.18\textwidth]{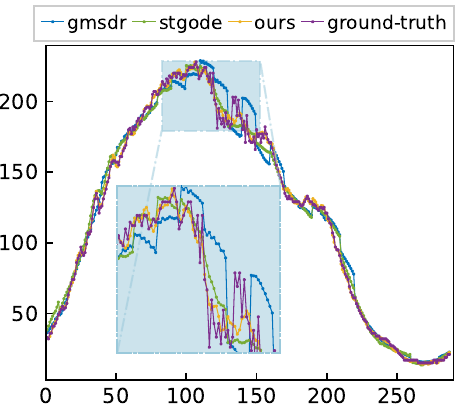}
}
\subfigure[node 198, from time step 505$\sim$793]{
    \centering
    \includegraphics[width=0.18\textwidth]{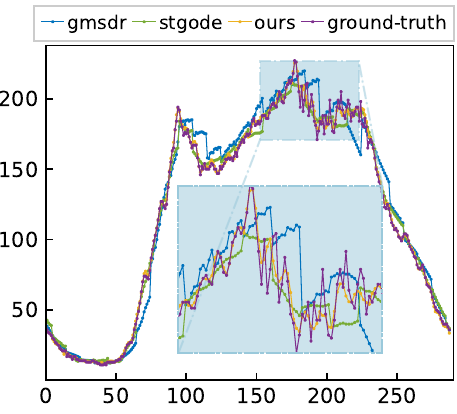}
}\vspace{-0.1in}
    \caption{Visualization of prediction results on PEMS04.}
    \vspace{-0.15in}
  \label{fig:pred_visual}
\end{figure}
\begin{table}[t]
  \centering
  \caption{Performance evaluation against data missing.}
  \vspace{-0.1in}
  \resizebox{0.49\textwidth}{!}{\begin{tabular}{c|ccc|ccc|ccc} 
  \hline
  \multirow{3}{*}{model} & \multicolumn{9}{c}{PEMS04}                                                                                \\ 
  \cline{2-10}
                         & \multicolumn{3}{c|}{missing 10\%} & \multicolumn{3}{c|}{missing 30\%} & \multicolumn{3}{c}{missing 50\%}  \\ 
  \cline{2-10}
                         & MAE   & RMSE  & MAPE              & MAE   & RMSE  & MAPE              & MAE   & RMSE  & MAPE              \\ 
  \hline
  STGODE                 & 23.97 & 35.41 & 19.13             & 45.02 & 59.48 & 29.54             & ~-~   & ~-~   & ~-~               \\
  GMSDR                  & 21.69 & 34.06 & 13.81             & 25.02 & 38.45 & 15.01             & 103.01& 131.64& 47.31                  \\
  \model\                 & 19.09 & 31.16 & 12.65             & 20.06& 32.94 & 13.11             & 20.66 & 33.74 & 13.58              \\
  \hline
  \end{tabular}}
  \label{tb:missing}
  \vspace{-0.15in}
  \end{table}

\begin{figure}
  \centering
    \vspace{-0.02in}
  \subfigure[On NYC Crime]{
      \centering
      \includegraphics[width=0.47\textwidth]{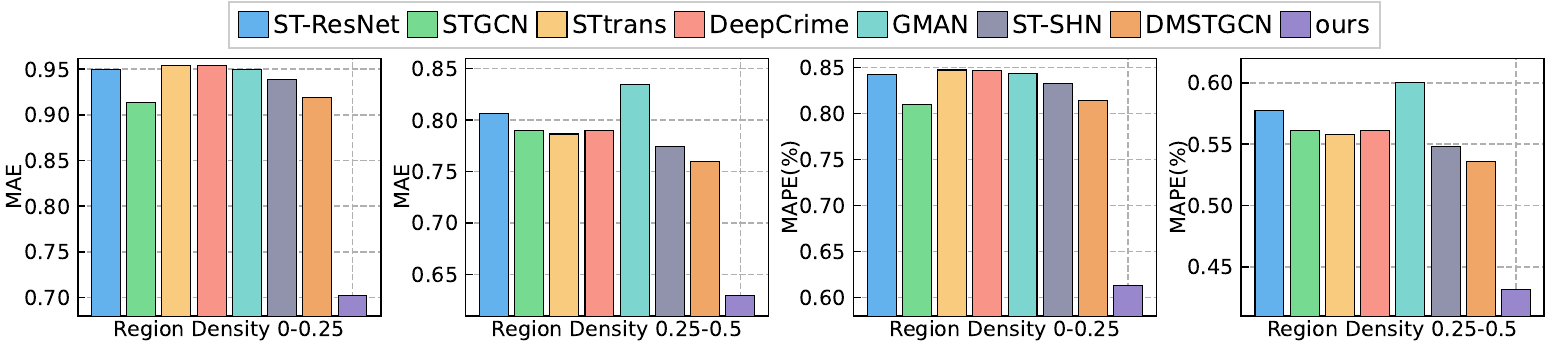}
  }\vspace{-3mm}
  \subfigure[On CHI Crime]{
    \centering
    \includegraphics[width=0.47\textwidth]{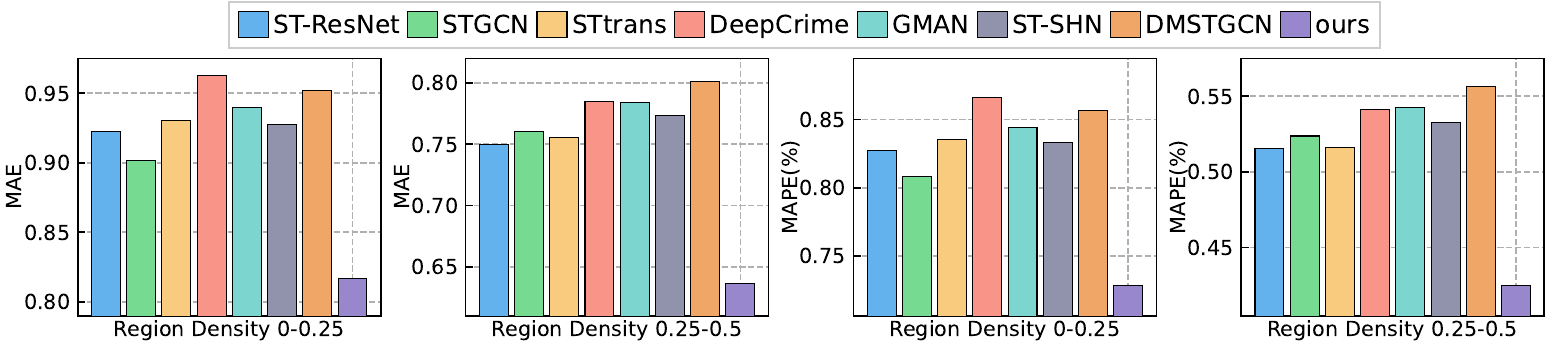}
  }
  \vspace{-4mm}
  \caption{Performance on sparse regions in crime prediction. }
  \vspace{-4mm}

  \label{fig:sparsity}
\end{figure}
  
\begin{figure*}[h]
      \centering
        
      \subfigure[On PEMS4]{
          \centering
          \includegraphics[width=0.31\textwidth]{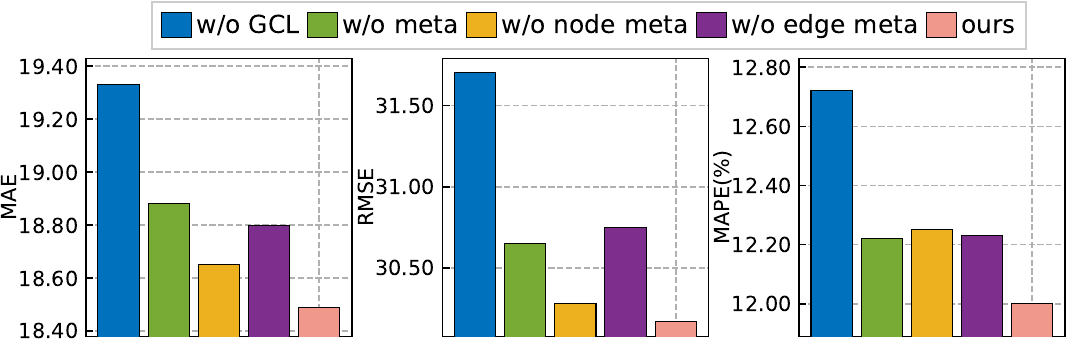}
          }
    \vspace{-0.1in}
      \subfigure[On PEMS7]{
        \centering
        \includegraphics[width=0.31\textwidth]{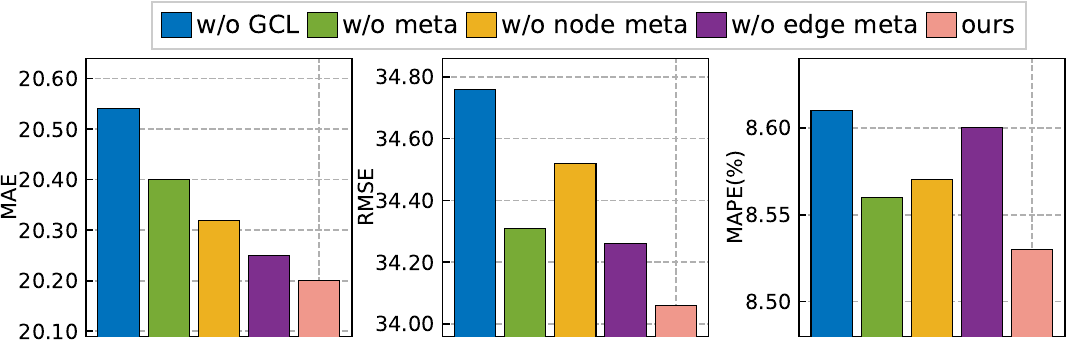}
      }
      \vspace{-0.1in}
      \subfigure[On PEMS8]{
        \centering
        \includegraphics[width=0.31\textwidth]{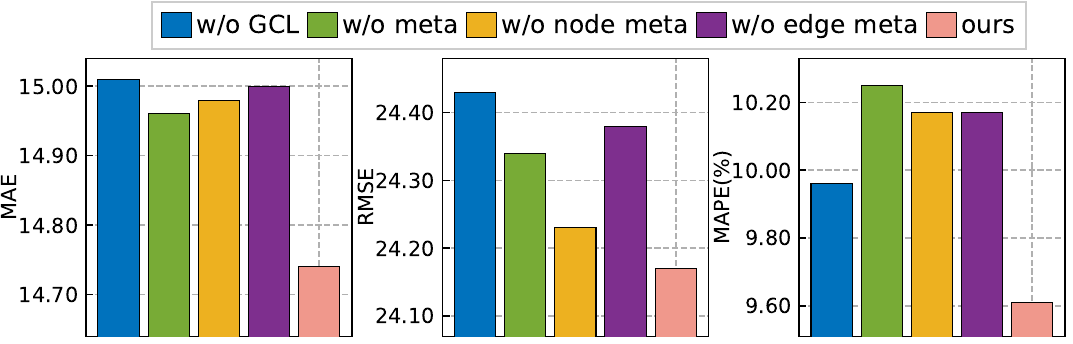}
      }
      \caption{Ablation experiments of our \model.}
      \vspace{-0.1in}
    
      \label{fig:ablation}
\end{figure*}
\begin{figure}
  \centering
    \includegraphics[width=0.47\textwidth]{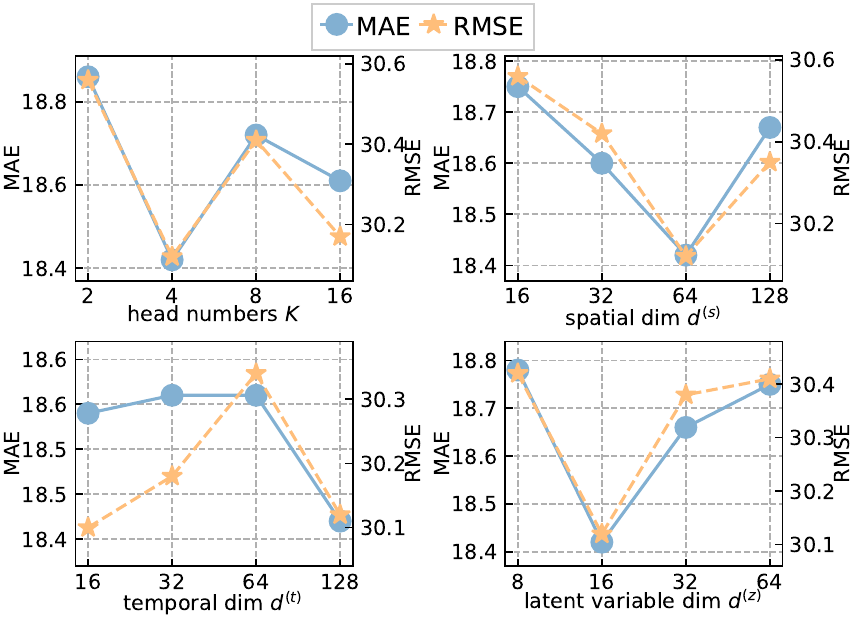}
    \vspace{-0.15in}
    \caption{Hyperparameter Investigation of  \model.}
    \vspace{-0.1in}
  \label{fig:para}
\end{figure}

\vspace{-0.05in}
\subsection{Overall Performance Comparison (RQ1)}


We present the performance comparison results on PEMS04, 07, and 08 datasets between the \model\ and state-of-the-art baselines in Table~\ref{tab:cmp1}. Additionally, we provide the comparison results for crime prediction in Table~\ref{tab:cmp2} to explore the effectiveness of our \model. In each dataset, we highlight the results of the best-performing model. Our findings can be summarized as follows: \\\vspace{-0.12in}

\noindent \textbf{Overall Superiority of \model.} 
The \model\ consistently achieves the best performance compared to different types of state-of-the-art baselines in all cases. This validates the effectiveness and superiority of our approach. We attribute these considerable improvements to two key factors: i) We integrate a meta view generator into the spatio-temporal graph contrastive learning framework, along with unified spatio-temporal GAT layers. This combination enhances the capability of encoding generalizable and robust spatio-temporal graph representations. ii) The view generator equipped with VAE-based meta networks automatically customizes optimal augmentation strategies for individual spatio-temporal graphs based on historical spatio-temporal contextual information. \\\vspace{-0.12in}

\noindent \textbf{Comparison with State-of-the-Arts.}
Although GNN-based models like FOGS, GMSDR, TAMP-S2GCNets, Z-GCNETs (for traffic), and DMSTGCN, ST-SHN, GMAN (for crime) are regarded as state-of-the-art solutions for spatio-temporal modeling, most of them rely on independently designed modules to capture spatial and temporal dependencies. However, this approach often leads to over-smoothing when multiple layers are stacked to improve representations. In comparison, our \model\ demonstrates significant improvements by employing a unified STG encoder and decoder with attention mechanisms. This unified approach allows us to learn global spatio-temporal dynamics with fewer layers, thanks to its enhanced representative capability. Moreover, when comparing our \model\ to attention-based methods such as DSANet, ASTGCN (for traffic), and DeepCrime, STtrans (for crime), we observe a performance gap. This gap highlights the enhanced representative ability of spatio-temporal meta contrastive learning framework, which enables our model to better capture the intricate customized spatio-temporal dynamics present in the data. \\\vspace{-0.12in}


\noindent \textbf{Visualization of Prediction Results.} 
To provide a more intuitive demonstration of \model's superiority over state-of-the-art baselines, we visualize the prediction results. In Figure~\ref{fig:pred_visual}, we present the prediction results of our \model\ alongside the ground-truth results and the results obtained by two competitive approaches, namely STG-ODE and GMSDR. Upon examining the visualization, we can observe that the prediction accuracy of our \model\ surpasses that of the other models, particularly when predicting traffic flow during instances of sharp changes or jitters. This improvement can be attributed to the fact that the spatio-temporal GAT encoder and decoder, trained using our designed contrastive learning paradigm, can effectively capture spatio-temporal dependencies. 



\vspace{-0.12in}
\subsection{Ablation Study (RQ2)}
To validate the effectiveness of the designed modules, we conduct ablation experiments on key components of our \model, namely the view generator with meta networks and the STG contrastive learning paradigm. The experimental results on traffic datasets are presented in Figure~\ref{fig:ablation}, and we make the following discoveries:\\\vspace{-0.12in}


\noindent (i) We remove the node-wise and edge-wise meta networks from the view generator to individually investigate their impact on the framework. This gives rise to the variants "w/o node meta" and "w/o edge meta". The results indicate that both node-wise and edge-wise meta networks contribute to improving the predictive performance independently. The node-wise meta networks extract spatio-temporal information from each graph and incorporate it into the generation of augmented views for nodes. On the other hand, the edge-wise meta networks learn task-relevant correlations and integrate them into the customized augmentation for edges. \\\vspace{-0.12in}


\noindent (ii) To confirm the effectiveness of the personalized view generator with meta networks, we design the variant "w/o meta" where the meta networks are replaced with randomly initialized optimizable parameters. We observe that the meta-knowledge enhanced view generators utilize the spatio-temporal latent correlations and inject spatio-temporal information into the framework, facilitating the acquisition of optimal augmentations. \\\vspace{-0.12in}


\noindent (iii) We conducted an additional experiment in which we removed the graph contrastive learning (GCL) framework from our \model\ and utilized a single original branch instead. This resulted in the creation of the variant "w/o GCL". During our analysis, we observed that the removal of the STG contrastive learning paradigm had a considerable negative impact on the performance of our \model. This finding highlights the crucial role played by the GCL framework in enhancing the model's effectiveness.




\vspace{-0.05in}
\subsection{Generalization and Robustness Study (RQ3)}
In this subsection, we demonstrate the generalization and robustness of our \model\ framework against the aforementioned challenges, specifically data missing and sparsity issues. \\\vspace{-0.12in}

\noindent\textbf{Performance \wrt\ Data Missing.}
As previously mentioned, data missing is a common challenge in real-world spatio-temporal scenarios, which can hinder the performance of advanced models. To assess the impact of data missing on our \model, we randomly drop the traffic volumes of nodes across the entire city independently, with missing proportions of 10\%, 30\%, and 50\% on the PEMS04 dataset. We compare the performance of our \model\ with two state-of-the-art approaches, namely STGODE and GMSDR. The comparison results are presented in Table~\ref{tb:missing}, where a "-" indicates that the method fails in that particular case. It can be observed that the predictive accuracy of the compared models significantly decreases with higher proportions of missing data. However, thanks to the designed contrastive learning paradigm, our \model\ can effectively adapt to the data missing scenario and encode more robust and generalizable STG representations. \\\vspace{-0.12in}


\noindent\textbf{Performance \wrt\ Data Sparsity.} 
To evaluate the performance of our \model\ in addressing data sparsity issues in real-world spatio-temporal prediction tasks, such as crime prediction and epidemic case prediction, we categorize the regions into four classes based on the historical density of crime signals in each region. These density classes are defined as "0-0.25", "0.25-0.5", "0.5-0.75", and "0.75-1.0". We compare the predictive results of our \model\ with baseline models specifically on regions with density classes of "0-0.25" and "0.25-0.5", as illustrated in Figure~\ref{fig:sparsity}. Significant performance gaps can be observed in all cases for the sparse regions. We attribute these improvements to the STG contrastive learning paradigm, which provides the STGNN with more supervised signals to enhance its representational capacity in this challenging and extreme scenario.


\begin{figure}[t]
  \centering
    
  \subfigure[Visualization of customized augmentations]{
      \centering
      
      \begin{minipage}[c]{0.48\textwidth} 
        \centering
        \includegraphics[width=0.48\textwidth]{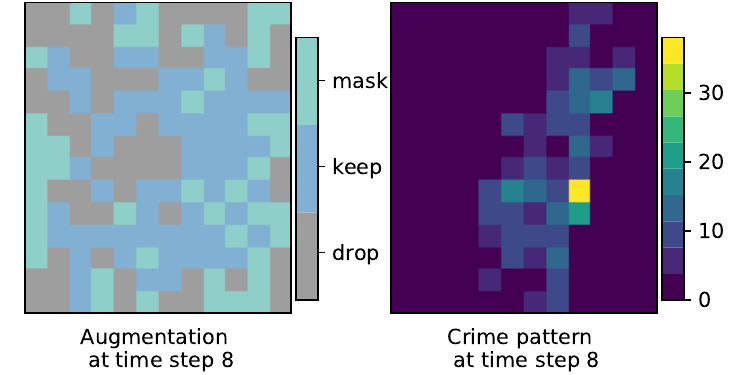}
      \includegraphics[width=0.48\textwidth]{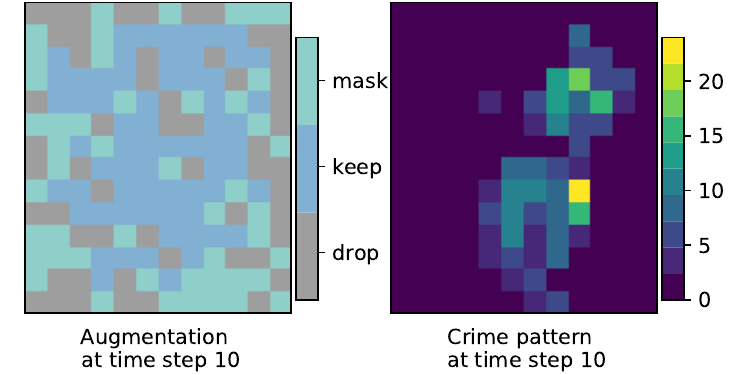}
      \includegraphics[width=0.48\textwidth]{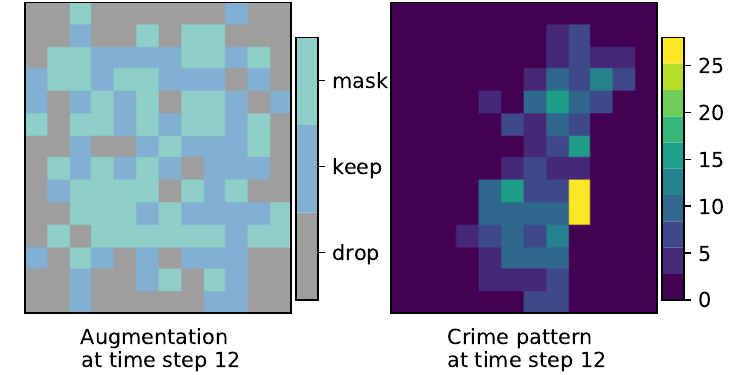}
      \includegraphics[width=0.48\textwidth]{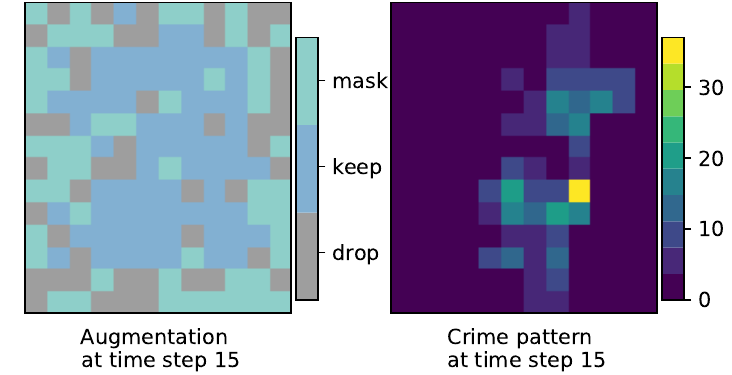}
        \end{minipage}
  }\vspace{-3mm}
  \subfigure[Semantics information from learned attention scores]{
    \centering
    \includegraphics[width=0.48\textwidth]{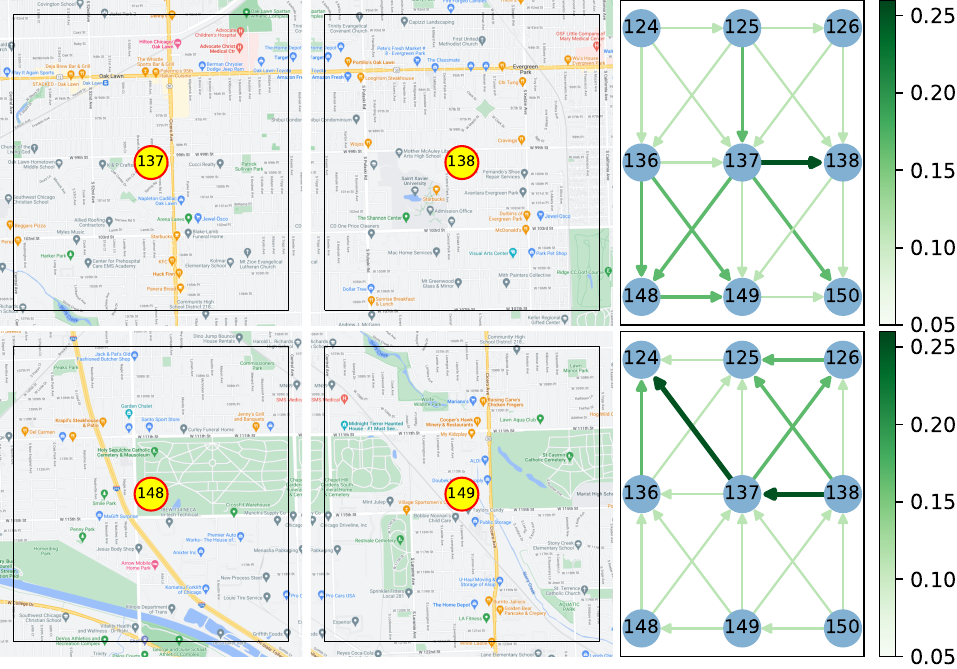}
  }
  \vspace{-0.2in}
  \caption{Case study of our \model\ Model. }
  \vspace{-0.2in}

  \label{fig:case}
\end{figure}

\vspace{-0.05in}
\subsection{Hyperparameter Investigation (RQ4)}
To investigate the influence of various hyperparameter settings, we conducted hyperparameter experiments by varying specific hyperparameters while keeping others at their default values. The experimental results on the PEMS04 dataset are presented in Figure~\ref{fig:para}. The following conclusions can be drawn:



(i) We search for head numbers, denoted as $K$, for the first spatio-temporal Graph Attention (GAT) layer in the spatio-temporal GAT encoder. We vary $K$ within the range of {2, $2^2$, $2^3$, $2^4$}. The results show that the best performance is achieved when $K = 2^2$. Interestingly, as we further increase the value of $K$, the prediction accuracy begins to somewhat deteriorate. This implies that a higher model representation capacity is not necessarily correlated with larger head numbers. (ii) We vary the spatial and temporal dimensions in the spatio-temporal GAT encoder. The search range for $d^{(s)}$ (spatial dimension) and $d^{(t)}$ (temporal dimension) is {$2^4$, $2^5$, $2^6$, $2^7$}, respectively. The results indicate that $d^{(s)} = 64$ is adequate to capture latent spatial dependencies in traffic patterns, whereas modeling temporal correlations requires $d^{(t)} = 128$. (iii) $d^{(z)}$ denotes the dimension of the latent variable in Equation~\ref{eq:paragen}, and our experimental search range is set to {$2^3$, $2^4$, $2^5$, $2^6$}. We observe that the best prediction accuracy is achieved with $d^{(z)} = 2^4$, and larger $d^{(z)}$ may introduce unexpected and task-irrelevant noise into the meta networks, thereby affecting the predictive performance negatively.


\vspace{-0.05in}
\subsection{Model Interpretation Case Study (RQ5)}
We investigate the model interpretation ability with a spatio-temporal GAT encoder enhanced by customized meta view generators. We explore two perspectives: (i) Whether the meta view generator constructs customized augmentations for different STGs based on their spatio-temporal patterns, and (ii) How the spatio-temporal GAT encoder captures spatio-temporal dynamics using contrastive learning. We visualize the customized augmentations and attention scores of randomly sampled STGs from the CHI crime datasets. \\\vspace{-0.12in}

\noindent \textbf{Visualization of Customized Augmentations.} 
In Figure~\ref{fig:case} (a), we present the visualization of customized regional augmentations for spatial graphs. Three different colors are used to represent three augmentation strategies, while the ground-truth crime records at different time steps are also shown. Our observations indicate that in areas with a high incidence of crime, the optimal augmentations tend to preserve the original data and apply drop and mask operations using the average value of crime records in other areas. This suggests that the designed meta networks effectively introduce spatio-temporal information into the learnable generation process, thereby filtering out task-irrelevant noise. Moreover, the visualization results demonstrate the diversity of customized augmentations for each STG, as evident in the distinct augmentation pattern at time step 12 compared to others.
\textbf{Semantics Learned with Attention Scores.} 
We visualize the learned attention scores of the trained spatio-temporal GAT encoder for the CHI crime dataset in Figure~\ref{fig:case} (b). The visualizations show strong correlations between regions (e.g., between region 137 and 138, and region 148 and 149) with similar urban functional properties, indicated by shared Point of Interest (POI) distributions. This suggests that the encoder effectively aggregates spatio-temporal semantics from neighboring regions, leading to accurate predictions. Overall, the results demonstrate the rationality and effectiveness of the trained spatio-temporal GAT encoder in capturing meaningful spatio-temporal dynamics.


\section{Related Work}

\subsection{DNNs for Spatio-Temporal Prediction}
Spatio-temporal prediction is crucial for various real-world applications, including traffic prediction~\cite{DCRNN, ST-MetaNet} and crime prediction~\cite{DeepCrime, ST-SHN}. With the advancements in deep learning techniques, researchers have employed Convolutional Neural Networks (CNNs) to capture spatial correlations in traffic flow~\cite{DCRNN, STGCN, GraphWaveNet}.  Attention mechanisms have been widely used in spatio-temporal traffic flow prediction to capture correlations in both time and space dimensions~\cite{ST-WA, STDN, GMAN}. In crime prediction, specific challenges such as data sparsity and skewed data distribution have led to the emergence of various approaches, including the use of hypergraph networks~\cite{ST-SHN} and self-supervised learning~\cite{ST-HSL} to address the unique characteristics of crime data. These advancements have significantly contributed to the progress of spatio-temporal prediction.

\subsection{Contrastive Learning On Graphs}
Contrastive learning has experienced significant advancements in recent years, emerging as a prominent component of self-supervised learning in various fields like computer vision~\cite{SimCLR} and natural language processing~\cite{CLEAR}. This learning approach has also demonstrated its efficacy in graph-structural data, offering powerful representation capabilities. By minimizing the contrastive loss, graph self-supervised learning effectively reduces the distance between positive sample pairs in the representation space while increasing the distance between negative sample pairs, thereby enhancing graph representations' robustness. Methods such as DGI~\cite{DGI} leverage both graph-level and node-level representations from the same input graph as positive sample pairs. They incorporate global representation information into local graph embeddings by maximizing mutual information. Another approach, MVGRL~\cite{MVGRL}, introduces a diffusion graph view in addition to the original view, maximizing mutual information to obtain resilient graph representations.


Graph contrastive learning relies on obtaining different views of the graph through graph data augmentation. In the study by GraphCL~\cite{GCLA}, four distinct graph-level data augmentation methods are proposed, highlighting the importance of augmentation strategies. Previous research has recognized the significance of finding optimal graph augmentations that can maximize the performance of contrastive learning. To enhance the effectiveness of graph augmentations, previous works have explored various approaches. Some studies have employed adaptive algorithms~\cite{JOAO}, which dynamically adjust the augmentation strategy based on the graph's characteristics or the learning progress.


\section{Conclusion}
\label{sec:conclusion}


In this study, we address several challenges in spatio-temporal prediction, including data quality and limitations of existing augmentations. To overcome these issues and generate robust and generalizable representations of spatio-temporal graphs (STG), we propose a novel framework called \model. Our framework incorporates personalized node- and edge-wise view generators with meta networks. This enables us to customize optimal augmentations for each STG, thereby enhancing the effectiveness of the contrastive learning paradigm. Additionally, we integrate spatio-temporal-aware information into the framework, further improving its performance. Furthermore, we introduce a spatio-temporal graph attention network encoder and a position-aware decoder within the contrastive learning paradigm. Extensive experiments demonstrate that our \model\ surpasses state-of-the-art approaches in terms of accuracy and robustness. This achievement validates the effectiveness of our proposed framework. In our future work, we aim to explore methods for enhancing the learnable view generation process. This may involve investigating denoising diffusion models or incorporating more explainable techniques to improve the quality and interpretability of the generated views.

\clearpage
\balance
\bibliographystyle{ACM-Reference-Format}
\bibliography{sample-base}



\end{document}